\documentclass{article}

\usepackage{array, multirow, booktabs}
\usepackage{graphicx}
 \usepackage[preprint]{neurips_2026}

\usepackage[utf8]{inputenc} 
\usepackage[T1]{fontenc}    
\usepackage{hyperref}       
\usepackage{url}            
\usepackage{booktabs}       
\usepackage{amsfonts}       
\usepackage{nicefrac}       
\usepackage{microtype}      
\usepackage{xcolor}         
\usepackage{amsmath}

\title{EVA: Evidence-seeking Visual Agent for Hallucination-Resistant Multimodal Reasoning}

\author{%
  Ting Lei \quad Yu Chen \quad Yang Liu\thanks{Corresponding author.} \\
  Wangxuan Institute of Computer Technology, Peking University \\
  \texttt{ting\_lei@pku.edu.cn} \quad
  \texttt{yu\_chen@stu.pku.edu.cn} \quad
  \texttt{yangliu@pku.edu.cn} \\
}

\begin{document}

\maketitle

\begin{abstract}

Hallucination remains a fundamental challenge in multimodal large language models (MLLMs), often stemming from unverified visual assumptions and reasoning processes that are weakly grounded in observable evidence. Existing approaches attempt to mitigate hallucination through self-reflection, multi-agent debate, or counterfactual verification. However, these methods remain largely opinion-driven: they lack explicit verification of the visual understanding process and may fail when multiple agents share the same incorrect perceptual assumptions, leading to consistent yet hallucinated conclusions.
A promising direction is to augment MLLMs with external visual tools (e.g., OCR and grounding models) to acquire verifiable evidence. However, effective tool-augmented reasoning remains challenging due to two key failure modes: unreliable tool invocation, where models select inappropriate tools or produce incorrect parameters, and evidence--reasoning inconsistency, where models ignore or contradict the acquired evidence when forming final predictions.
In this work, we propose EVA, an Evidence-seeking Visual Agent that reframes multimodal reasoning as an explicit process of evidence acquisition and verification. At the core of EVA is a unified evidence-grounded execution pipeline designed to address the above challenges through two tightly coupled mechanisms. First, we introduce \textit{Iterative Tool Replanning}, which dynamically refines tool selection and parameters based on intermediate observations, improving the reliability of tool usage. Second, we develop \textit{Evidence Consistency Verification}, an answer validation mechanism that improves alignment between the final prediction and the collected evidence, mitigating evidence--reasoning inconsistencies.
In addition, to balance robustness and efficiency, EVA incorporates \textit{Selective Evidence Routing}, an adaptive fast--slow reasoning strategy that determines when external evidence is necessary. Visually straightforward queries are handled via direct reasoning, while complex or high-risk cases trigger tool-based evidence acquisition.
By jointly addressing how to reliably acquire evidence and how to use it consistently, EVA transforms multimodal reasoning from passive answer validation into explicit evidence-backed decision making. Experimental results demonstrate that EVA achieves significantly improved reasoning reliability over direct VLM inference and multi-agent methods, while maintaining efficiency through selective tool invocation. These findings highlight structured and adaptive evidence utilization as a practical paradigm for hallucination-resistant multimodal reasoning. Code is available at \url{https://github.com/oceanflowlab/EVA}.

\end{abstract}

\section{Introduction}

Multimodal large language models (MLLMs) have recently demonstrated strong capabilities in visual understanding and reasoning. However, hallucination remains a fundamental obstacle to their reliability, especially in multimodal settings where reasoning depends on both visual perception and language generation. In many cases, models produce plausible but incorrect answers grounded in unsupported visual assumptions, or fail to verify whether their reasoning is consistent with observable evidence. This issue is particularly critical for tasks requiring fine-grained perception, such as reading text in images, identifying subtle attributes, or interpreting structured visual content.

\begin{figure}[t]
\centering
\includegraphics[width=\linewidth]{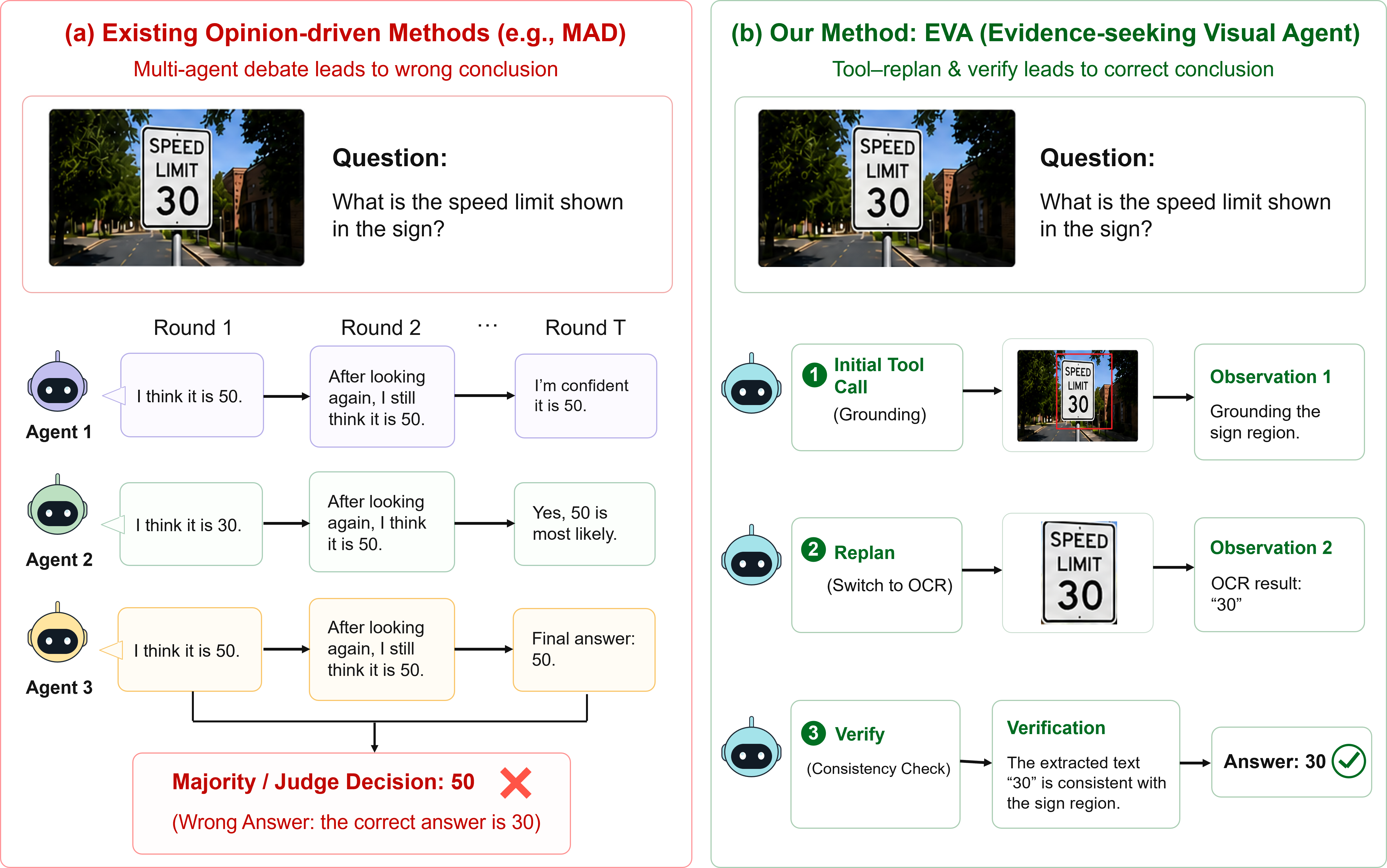}
\caption{\textbf{Opinion-driven vs. evidence-driven multimodal reasoning.}
(a) Existing methods rely on multi-agent debate or iterative reasoning, where agents may converge to an incorrect answer due to shared perceptual errors.
(b) EVA actively seeks and verifies visual evidence through tool use and replanning, ensuring that the final answer is grounded in reliable observations.
}
\label{fig:teaser}
\vspace{-1em}
\end{figure}

A growing body of work attempts to mitigate hallucination by improving reasoning robustness. Representative approaches include self-reflection, multi-agent debate (MAD), and counterfactual verification. These methods aim to identify unreliable outputs through iterative reasoning, cross-agent interaction, or hypothetical testing. While promising, they remain largely \emph{opinion-driven}: reasoning is conducted primarily in the language space, and correctness is often inferred from agreement, consistency, or confidence rather than explicit grounding in visual evidence. As illustrated in Figure~\ref{fig:teaser}(a), such approaches may lead to consistent yet incorrect conclusions, even when multiple agents participate in deliberation. In particular, when different reasoning paths share the same incorrect perceptual assumptions, agreement among agents does not guarantee correctness.

We argue that hallucination in multimodal reasoning is not merely a failure of reasoning, but more fundamentally a failure of \emph{evidence acquisition and verification}. Reliable reasoning should be grounded in explicitly acquired, task-relevant visual evidence, rather than inferred from internal deliberation or inter-agent consensus.

A natural solution is to augment MLLMs with external visual tools (e.g., OCR and grounding models) to acquire verifiable evidence. However, effective tool-augmented reasoning introduces new challenges. First, models often suffer from \emph{tool invocation unreliability}, where inappropriate tools are selected or incorrect parameters are produced, leading to noisy or irrelevant observations. Second, even when correct evidence is obtained, models may exhibit \emph{evidence--reasoning inconsistency}, ignoring or contradicting the collected evidence when forming final predictions. These issues reveal that simply providing tools is insufficient; reliable multimodal reasoning requires both accurate evidence acquisition and consistent evidence utilization.

To address these challenges, we propose EVA, an Evidence-seeking Visual Agent that reframes multimodal reasoning as an explicit evidence-driven process. At the core of EVA is a unified evidence-grounded execution pipeline composed of two tightly coupled mechanisms. First, we introduce \textit{Iterative Tool Replanning}, which dynamically refines tool selection and parameters based on intermediate observations, improving the reliability of tool invocation. Second, we develop \textit{Evidence Consistency Verification}, an answer validation mechanism that enforces alignment between the final prediction and the collected evidence, mitigating evidence--reasoning inconsistencies. As shown in Figure~\ref{fig:teaser}(b), this replan-and-verify process enables the model to correct earlier decisions and produce evidence-consistent answers.

In addition, to balance robustness and efficiency, EVA incorporates \textit{Selective Evidence Routing}, an adaptive fast--slow reasoning strategy that determines when external evidence is necessary. Visually straightforward queries are handled via direct reasoning, while complex or high-risk cases trigger tool-based evidence acquisition (e.g., OCR, object grounding, and chart parsing). This design avoids unnecessary tool usage while preserving robustness on challenging inputs.

By jointly addressing how to reliably acquire evidence and how to use it consistently, EVA transforms multimodal reasoning from passive validation of generated answers into explicit evidence-backed decision making.

Our contributions are threefold:
(1) We reformulate hallucination-resistant multimodal reasoning as an evidence acquisition and verification problem, highlighting the limitations of opinion-driven validation in multimodal settings.
(2) We propose EVA, a unified evidence-seeking visual agent built upon an evidence-grounded execution pipeline, featuring \textit{Iterative Tool Replanning}, \textit{Evidence Consistency Verification}, and \textit{Selective Evidence Routing}.
(3) Experiments demonstrate that EVA improves reasoning reliability while maintaining efficiency through selective evidence utilization.

\section{Related Work}

\subsection{Multimodal Reasoning for Hallucination Removal}

Early efforts in multimodal reasoning focused on aligning visual and textual representations, leading to the development of multimodal large language models (MLLMs)~\cite{liu2023improvedllava,chen2024internvl} and the adoption of in-context learning (ICL)~\cite{baldassini2024ICL} and chain-of-thought (CoT)~\cite{wei2022CoT} reasoning. These techniques enable models to decompose complex tasks into interpretable intermediate steps, improving both performance and transparency. The extension of CoT to multimodal settings, often referred to as multimodal chain-of-thought (MCoT), has led to diverse reasoning paradigms, including linear~\cite{zhang2023MCoT} and graph-based~\cite{besta2024graph-of-thought} structures, as well as modality-specific designs such as MVoT~\cite{li2025imagine-MVOT}, and Video-of-Thought~\cite{fei2024video-of-thought}.
To further enhance reasoning robustness, multi-agent systems (MASs)~\cite{wooldridge2009MAS} have been introduced to leverage collective intelligence across multiple agents. In particular, multi-agent debate (MAD) frameworks~\cite{zheng2024picture,wang2025model,lin2024interpreting} aim to mitigate hallucination by aggregating diverse reasoning paths through interaction and consensus. Recent work~\cite{liang2025MUG} further reframes MAD as a process of detecting undercover agents by employing multimodal counterfactual tests. 

Despite these advances, existing approaches remain largely confined to language-space reasoning and implicit verification signals, such as agreement, consistency, or counterfactual plausibility. As a result, they may still fail when multiple agents share the same incorrect visual assumptions, leading to consistent but unsupported conclusions. Moreover, these methods lack mechanisms for selectively acquiring task-relevant visual evidence, and therefore cannot adaptively determine when additional perceptual grounding is required. In contrast, our approach introduces an adaptive fast–slow reasoning mechanism that explicitly triggers evidence acquisition when necessary, enabling more reliable multimodal reasoning grounded in observable evidence.

\subsection{Tool-Augmented Multimodal Reasoning}

Recent work has explored augmenting MLLMs with external tools to improve visual reasoning capabilities. Early efforts focused on building infrastructure and datasets for tool integration. For instance, LLaVA-Plus~\cite{liu2024Llava-plus} introduced a dedicated tool server, while CogCoM~\cite{qi2024cogcom} and TACO~\cite{liu2024taco} constructed large-scale training data with structured manipulation traces and tool-augmented reasoning demonstrations. These approaches aim to teach models how to interact with tools through supervised learning.
Subsequent research has investigated different paradigms for tool use. One prominent direction~\cite{zhao2025pyvision,zhang2025thyme} trains models to generate executable code for complex reasoning tasks, enabling flexible and expressive interactions with external environments. Another line of work~\cite{wang2025vgr,wang2025divide,wang2025pixel,zhu2025active,zheng2025deepeyes,song2026adareasoner} focuses on lightweight, atomic visual tools, such as zoom-in or region selection, to enhance perception in a more structured and controllable manner.

However, existing tool-augmented approaches typically rely on extensive data curation and model fine-tuning, resulting in high training costs and limited scalability. Furthermore, their tool-use policies are often brittle, exhibiting poor generalization to unseen task distributions beyond the training scope. In contrast, our approach treats tool use as an adaptive inference-time process, guided by explicit evidence acquisition and verification. By combining selective fast–slow reasoning with a replan-and-verify execution pipeline, EVA enables flexible and reliable tool use without requiring exhaustive supervision or fixed tool policies.

\section{Method}

\subsection{Problem Formulation}

We study hallucination-resistant multimodal reasoning, where a model answers a question grounded in an input image using verifiable visual evidence. Given an image $I$ and a question $q$, the goal is to predict an answer $a$ based on task-relevant evidence $E$ extracted from the image:
\begin{equation}
a = f(q, I, E), \quad \text{s.t. } E \subseteq I \text{ and } E \Rightarrow a.
\end{equation}

Here, $E$ denotes observable evidence such as text regions, objects, attributes, or spatial relations. The condition $E \Rightarrow a$ requires the answer to be supported by explicit visual evidence rather than implicit visual assumptions.

However, reliable evidence-grounded reasoning remains challenging for tool-augmented MLLMs. We identify two key issues: (1) \textbf{Tool Invocation Reliability:} The model may select inappropriate tools or generate suboptimal tool parameters. (2) \textbf{Evidence--Reasoning Inconsistency:} The model may ignore or contradict the acquired evidence during answer generation.
To address these challenges, we formulate multimodal reasoning as an evidence-grounded decision process that iteratively acquires evidence and verifies answer consistency.

\subsection{Overview of EVA}

We propose EVA, an Evidence-seeking Visual Agent that reframes multimodal reasoning as explicit evidence acquisition and verification. EVA is built around an evidence-grounded execution pipeline designed to address the two challenges above.
The pipeline contains two core mechanisms. First, \textit{Iterative Tool Replanning} improves tool invocation reliability by refining tool selection and parameters based on intermediate observations. Rather than relying on a single-pass tool plan, EVA allows the model to inspect obtained evidence and replan subsequent tool usage. Second, \textit{Evidence Consistency Verification} addresses evidence--reasoning inconsistency by checking whether the synthesized answer is fully supported by the collected evidence and revising unsupported conclusions.
In addition, EVA incorporates \textit{Selective Evidence Routing}, an adaptive fast--slow reasoning mechanism that determines whether external evidence is necessary for a given query. Since many visually simple questions can be answered directly, this routing mechanism avoids unnecessary tool invocation while preserving robustness for complex or high-risk cases.

\begin{figure*}[t]
    \centering
    \includegraphics[width=\textwidth]{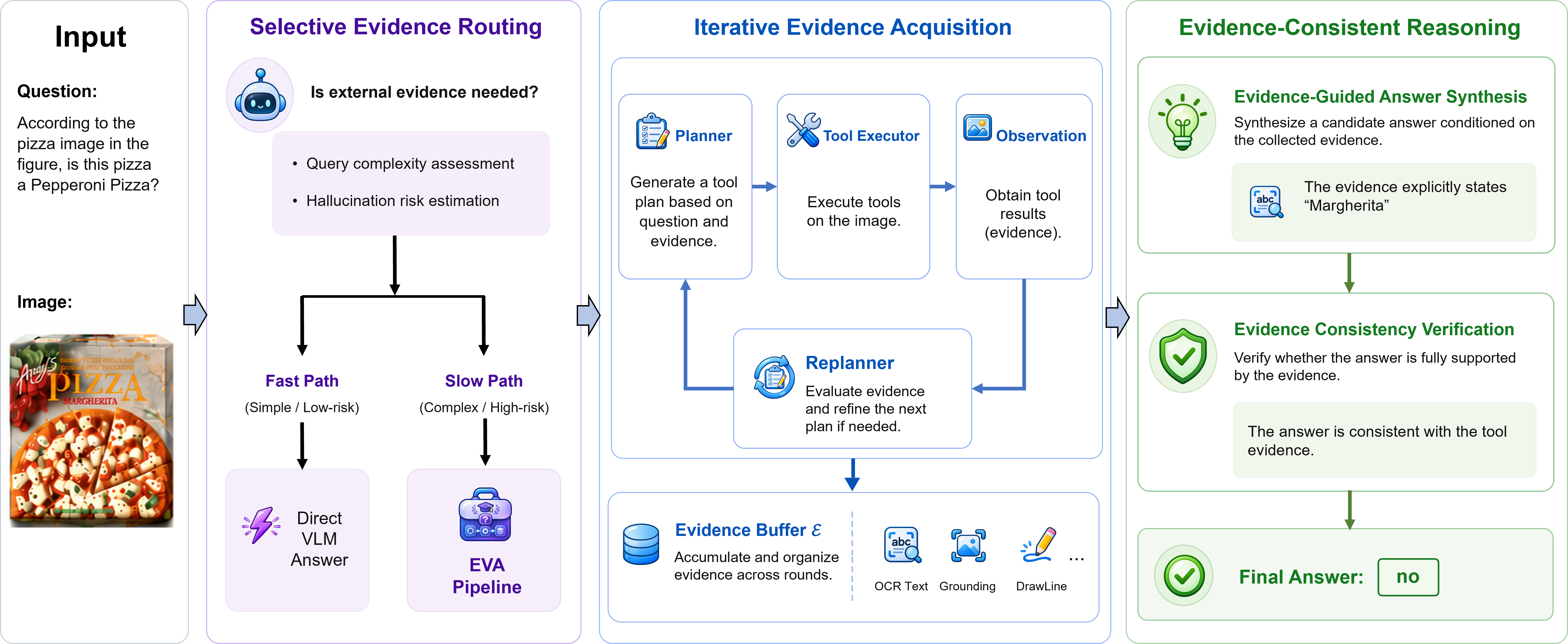}
    \caption{
    Overview of EVA.
    EVA first performs Selective Evidence Routing to determine whether external evidence is required.
    For complex queries, EVA activates an evidence-grounded execution pipeline with iterative tool replanning, evidence-guided answer synthesis, and evidence consistency verification.
    }
    \label{fig:eva}
    \vspace{-1em}

\end{figure*}

\subsection{Evidence-Grounded Execution Pipeline}

EVA's core execution pipeline targets reliable evidence acquisition and evidence-consistent answer generation. Given an image-question pair $(I, q)$, EVA first produces an initial tool-use plan:
\begin{equation}
(\mathcal{P}^{(0)}, \mathcal{T}^{(0)}) = \pi_{\theta}(I, q),
\end{equation}
where $\mathcal{P}^{(0)}$ denotes the initial reasoning plan and $\mathcal{T}^{(0)} = {t_1^{(0)}, \dots, t_k^{(0)}}$ denotes the corresponding tool calls, including tool types and parameters.

Each tool call is executed on the image to obtain visual evidence:
\begin{equation}
E_i^{(0)} = t_i^{(0)}(I), \quad E^{(0)} = {E_1^{(0)}, \dots, E_k^{(0)}}.
\end{equation}

However, a single tool invocation round can be unreliable. The initial plan may use an unsuitable tool, localize the wrong image region, apply incorrect OCR settings, or extract evidence that is insufficient for answering the question. EVA therefore introduces an iterative replanning process.

\paragraph{Iterative Tool Replanning.}

At each round $r$, EVA evaluates the accumulated evidence $E^{(r)}$ with respect to the question $q$ and updates the tool plan when the evidence is incomplete, irrelevant, or uncertain:
\begin{equation}
(\mathcal{P}^{(r+1)}, \mathcal{T}^{(r+1)}) = \pi_{\theta}(I, q, E^{(r)}).
\end{equation}

The new tool calls are then executed to acquire additional evidence:
\begin{equation}
\Delta E^{(r+1)} = \mathcal{T}^{(r+1)}(I), \quad
E^{(r+1)} = E^{(r)} \cup \Delta E^{(r+1)}.
\end{equation}

This mechanism allows EVA to dynamically adjust its evidence-seeking behavior. For example, if object grounding fails to resolve a queried entity, EVA may switch to OCR, or refine the target description depending on the intermediate observation. In this way, Iterative Tool Replanning directly improves tool invocation reliability by correcting suboptimal tool choices and parameters during execution.

\paragraph{Evidence-Guided Answer Synthesis.}

After evidence acquisition, EVA synthesizes a candidate answer from the collected evidence:
\begin{equation}
a_{\text{syn}} = g_{\theta}(q, E).
\end{equation}

Unlike direct MLLM inference, the synthesis step conditions the answer primarily on explicit evidence rather than implicit visual assumptions. The model is encouraged to aggregate relevant observations and produce an answer that is traceable to the acquired evidence.

\paragraph{Evidence Consistency Verification.}

Even when the evidence is sufficient, the synthesized answer may still be inconsistent with the tool results. To address this, EVA performs a final verification step:
\begin{equation}
a = h_{\theta}(q, E, a_{\text{syn}}).
\end{equation}

The verifier checks whether the candidate answer is entailed by the collected evidence. If the answer ignores relevant observations, contradicts tool outputs, or relies on unsupported assumptions, the verifier revises the prediction or rejects the unsupported reasoning path. This stage explicitly enforces evidence--reasoning consistency and reduces hallucinated conclusions caused by ungrounded answer synthesis.

\subsection{Selective Evidence Routing}

Although external tools improve reliability, invoking them for every query is inefficient and sometimes unnecessary. Many visually straightforward questions can be answered accurately through direct visual reasoning, while complex or high-risk queries require explicit evidence acquisition.

To balance efficiency and robustness, EVA introduces \textit{Selective Evidence Routing}. Before activating the full evidence-grounded execution pipeline, EVA predicts whether external evidence is needed:
\begin{equation}
z = \mathbb{I} [\text{need\_evidence}(I, q)].
\end{equation}

If $z=0$, EVA follows a fast path and directly answers the query:
\begin{equation}
a = f_{\theta}(I, q).
\end{equation}

If $z=1$, EVA follows the slow path and activates the full evidence-grounded pipeline:
\begin{equation}
a = h_{\theta}(q, E, g_{\theta}(q, E)).
\end{equation}

This adaptive strategy treats tool use as a targeted intervention rather than a default operation. It avoids unnecessary tool calls for simple cases, while routing complex, ambiguous, or hallucination-prone queries to the more reliable replan-and-verify pipeline.

\section{Experiments}

\subsection{Experimental Setup}


\paragraph{Evaluated Benchmarks.}
We adopt the VLMEvalKit~\cite{duan2024vlmevalkit} framework for fair evaluation. We evaluate EVA on four benchmark datasets that cover diverse aspects of multimodal reasoning:

\textbf{General Reasoning Benchmarks.}
We adopt MMStar~\cite{chen2024MMStar} and MMMU~\cite{yue2023mmmu} as comprehensive benchmarks for evaluating overall multimodal reasoning ability. MMStar consists of 1,500 challenging visual questions spanning multiple domains, while MMMU contains 11,500 college-level multimodal problems that require advanced cross-disciplinary reasoning.

\textbf{Hallucination Detection Benchmarks.}
We further evaluate hallucination robustness using HallusionBench~\cite{guan2024hallusionbench} and POPE~\cite{li2023POPE}. HallusionBench includes 346 carefully designed samples that induce hallucinations through misleading visual cues. POPE contains 3,000 binary (yes/no) questions aimed at systematically evaluating object hallucination in image descriptions.

\paragraph{Baselines.}
We compare EVA against a diverse set of baseline methods to comprehensively benchmark its performance:
\textbf{(1) Single-Agent Models.}
These include both open-source and closed-source multimodal models. Open-source baselines include DeepSeek-VL-7B~\cite{lu2024deepseekvl}, LLaVA-OneVision-7B~\cite{li2024LLaVA-OneVision}, LLaVA-NEXT-Llama3-8B~\cite{liu2024llavanext}, LLaVA-v1.5-13B~\cite{liu2023llava}, InternVL2-8B~\cite{chen2024internvl}, InternVL3-14B~\cite{zhu2025internvl3}, and Qwen2.5-VL~\cite{Qwen2.5-VL}. Closed-source baselines include Gemini-1.5-Pro~\cite{team2024gemini}, GPT-4o-mini~\cite{hurst2024gpt-4o}, GPT-4V~\cite{openai2023gpt4v}, and Claude-3.5-Sonnet~\cite{anthropic2024claude}. These models integrate visual and textual inputs using different architectural and training strategies to enhance multimodal understanding.
\textbf{(2) Self-Refine~\cite{madaan2023self-refine}.}
Self-Refine is an approach that improves initial outputs of large language models through iterative self-feedback and refinement.
\textbf{(3) Multi-Agent Debate (MAD)~\cite{liang2024MAD}.}
MAD represents a representative multi-agent paradigm, where multiple agents engage in iterative debate and reach final decisions via majority voting or a centralized judge. 
\textbf{(4) Multi-agent Undercover Gaming (MUG)~\cite{liang2025MUG}.}
MUG reframes multi-agent debate as a process of identifying “undercover” agents (i.e., hallucinating agents) via multimodal counterfactual testing, where modified visual evidence is introduced to verify agents’ consistency against ground truth.

\paragraph{Controlled Tool-Use Baselines.}
We introduce four controlled baselines using the same backbone, tools, and evaluation protocol as EVA to isolate the effects of tool access and inference computation. \textbf{OneShot} performs a single round of tool use without replanning or verification. \textbf{ReAct} iteratively selects tools based on intermediate observations without explicit evidence verification. \textbf{FixedSecondRound} always executes a second tool-use round regardless of evidence sufficiency. \textbf{SameEvidenceSelfRefine} refines its answer using exactly the same evidence acquired by EVA, isolating the effect of structured reasoning from evidence access.

\begin{table*}[t]
\centering
\resizebox{\textwidth}{!}{
\begin{tabular}{l|cc|cccc|cccc}
\toprule
\multirow{2}{*}{Model} & \multicolumn{1}{c}{MMMU} & \multicolumn{1}{c|}{MMStar} & \multicolumn{4}{c|}{HallusionBench} & \multicolumn{4}{c}{POPE} \\
 & Acc. &  Acc. & aAcc. & fAcc. & qAcc. & Avg. & Acc. & Pre. & Rec. & F1 \\
\midrule
DeepSeek-VL-7B & 38.3 & 40.5 & 53.9 & 24.9 & 24.6 & 34.5 & 86.8 & 94.3 & 78.4 & 85.6 \\
LLaVA-OneVision-7B & 47.9 & 61.9 & 48.4 & 21.4 & 25.1 & 31.6 & 87.9 & 96.7 & 78.5 & 86.6 \\
LLaVA-Next-Llama3-8B & 43.1 & 43.9 & 52.3 & 25.7 & 21.3 & 33.1 & 87.1 & 88.1 & 95.1 & 87.1 \\
LLaVA-v1.5-13B & 37.0 & 34.3 & 45.3 & 34.9 & 11.0 & 24.5 & 87.5 & 91.4 & 82.3 & 85.9 \\
InternVL2-8B & 50.2 & 61.5 & 63.9 & 35.0 & 36.0 & 45.0 & 86.0 & 96.6 & 74.7 & 84.2 \\
InternVL2-26B & 50.7 & 61.0 & 67.9 & 44.8 & 41.8 & 51.5 & 87.7 & 96.4 & 78.3 & 86.4 \\
Gemini-1.5-Pro & 60.6 & 59.1 & 63.0 & 36.1 & 37.6 & 45.6 & 88.6 & 93.0 & 83.9 & 88.2 \\
GPT4o mini (detail-high) & 57.3 & 46.5 & 61.9 & 41.3 & 34.9 & 46.1 & 84.2 & 95.5 & 71.7 & 81.9 \\
GPT-4v (detail-high) & 53.8 & 43.9 & 65.8 & 38.4 & 35.2 & 46.5 & 83.9 & 93.9 & 72.5 & 81.8 \\
Claude3.5-Sonnet & 65.9 & 62.2 & 66.4 & 41.6 & 41.8 & 49.9 & 78.7 & 97.1 & 59.2 & 73.6 \\
\midrule
Qwen2.5VL-7B & 45.0 & 61.2 & 64.8 & 34.9 & 39.6 & 46.4 & 87.4 & 96.8 & 77.3 & 85.9 \\
Qwen2.5VL-7B (Self-Refine) & 45.8 & 61.5 & 67.3 & 38.2 & 40.9 & 48.8 & 85.9 & 97.4 & 73.8 & 84.0 \\
Qwen2.5VL-7B (MAD-Vote) & 44.7 & 57.4 & 56.4 & 26.9 & 30.1 & 37.8 & 80.0 & 80.4 & 68.7 & 74.1 \\
Qwen2.5VL-7B (MAD-Judge) & 47.4 & 62.3 & 64.5 & 43.2 & 42.9 & 50.2 & 85.2 & 94.4 & 74.9 & 83.5 \\
Qwen2.5VL-7B (MUG) & 50.3 & 63.8 & 69.4 & 43.9 & \textbf{47.9} & 53.8 & \textbf{88.4} & \textbf{95.6} & {80.5} & \textbf{87.4} \\
Qwen2.5VL-7B (EVA(Ours)) & \textbf{51.4} & \textbf{64.3} & \textbf{70.4} & \textbf{45.4} & 47.3 & \textbf{54.4} & 88.1 & 94.1 & \textbf{81.3} & 87.3 \\
\midrule
InternVL3-14B & 59.8 & 68.7 & 69.8 & 47.7 & 47.7 & 55.1 & 89.3 & 92.1 & 86.5 & 89.5 \\
InternVL3-14B (Self-Refine) & 45.9 & 61.2 & 70.3 & 46.5 & 45.5 & 54.1 & 88.6 & 91.9 & 86.7 & 89.2 \\
InternVL3-14B (MAD-Vote) & 55.2 & 62.9 & 70.5 & 48.6 & 46.4 & 55.2 & 89.4 & 94.2 & 83.9 & 88.8 \\
InternVL3-14B (MAD-Judge) & 60.2 & 68.9 & 72.5 & 47.9 & 48.9 & 56.4 & 89.6 & 94.1 & 87.3 & 90.6 \\
InternVL3-14B (MUG) & 60.7 & \textbf{69.1} & \textbf{73.3} & {51.2} & 49.5 & 58.0 & \textbf{90.1} & 94.1 & \textbf{88.2} & \textbf{91.1} \\
InternVL3-14B (EVA(Ours)) & \textbf{62.7} & \textbf{69.1} & {72.2} & \textbf{52.3} & \textbf{51.9} & \textbf{58.8} & 89.5 & \textbf{94.2} & 84.2 & 88.9 \\
\bottomrule
\end{tabular}
}
\caption{Performance comparison across MMMU, MMStar, HallusionBench, and POPE benchmarks. Acc.: accuracy, aAcc.: average accuracy, fAcc.: unique figure accuracy, qAcc.: unique question accuracy, Avg.: average score of the three metrics. Pre.: precision, Rec.: recall.}
\label{tab:main_results}
\end{table*}

\subsection{Main Results}

Table~\ref{tab:main_results} summarizes the comparison between EVA and existing multimodal reasoning approaches across MMMU, MMStar, HallusionBench, and POPE. Overall, EVA consistently improves both reasoning capability and hallucination resistance across different model backbones, demonstrating the effectiveness of evidence-grounded reasoning.

\paragraph{EVA consistently improves both small and large MLLMs.}
Integrating EVA leads to stable gains on both Qwen2.5VL-7B and InternVL3-14B across nearly all benchmarks. On Qwen2.5VL-7B, EVA improves the baseline from 61.2\% to 64.3\% on MMStar, and from 46.4\% to 54.4\% on HallusionBench. Similar improvements are observed on InternVL3-14B, where EVA achieves 69.1\%, and 58.8\%, respectively. Notably, EVA built on Qwen2.5VL-7B surpasses several proprietary models such as GPT-4v and Claude3.5-Sonnet on HallusionBench, suggesting that structured evidence-grounded reasoning can effectively compensate for limitations in model scale.

\paragraph{EVA achieves stronger hallucination resistance than self-reflection and multi-agent methods.}
Compared with Self-Refine, MAD-Vote, MAD-Judge, and MUG, EVA achieves the best or near-best performance across most hallucination-related metrics. On HallusionBench, EVA obtains the highest average scores of 54.4\% and 58.8\% on the two backbones respectively, while also achieving strong figure and question accuracy. 
Compared with self-reflection and multi-agent approaches, EVA directly improves the reliability of evidence acquisition and evidence utilization instead of relying solely on answer refinement or opinion aggregation. While multiple agents may still converge to the same incorrect perceptual assumption, EVA explicitly verifies visual evidence through Iterative Tool Replanning and Evidence Consistency Verification, leading to more reliable multimodal reasoning.

\subsection{Behavior Analysis of EVA}
\label{sec:behavior_analysis}

To better understand EVA beyond aggregate benchmark performance, we analyze its tool usage behavior, iterative replanning dynamics, and performance--efficiency trade-off on HallusionBench~\cite{guan2024hallusionbench}.

\subsubsection{Tool Usage Behavior}

We first analyze EVA's tool invocation behavior.
EVA activates external tools on only 19.87\% of samples, with an average of 0.323 tool calls per sample, indicating that \textit{Selective Evidence Routing} invokes tools conservatively rather than relying on external evidence by default.
Most queries are resolved through direct multimodal reasoning, while tool usage is primarily triggered for perception-intensive cases.

We further examine EVA's multi-round execution dynamics, where the tool invocation pattern also reveals the adaptive behavior of EVA.
The tool-call rate increases from 20.34\% in the first round to 56.61\% in the second round, indicating that EVA progressively refines earlier observations through targeted tool usage once additional evidence becomes necessary.
This behavior aligns with the design objective of \textit{Iterative Tool Replanning}, where intermediate observations guide subsequent tool selection and parameter refinement.

Interestingly, tool-triggered samples achieve higher accuracy (76.72\%) than no-tool samples (68.37\%), despite generally corresponding to more difficult perception-intensive queries.
This suggests that EVA effectively identifies cases where external evidence is beneficial and improves reasoning reliability through targeted evidence acquisition.

Figure~\ref{fig:tool_distribution} further illustrates the distribution of external tool usage in EVA.
OCR is the most frequently invoked tool, followed by chart-understanding and grounding-related tools.
This distribution suggests that many challenging multimodal reasoning cases require fine-grained perception capabilities, particularly for text recognition and structured visual understanding.

Finally, EVA demonstrates highly reliable tool execution in practice, with failures~(undefined tool name output) observed in only 0.11\% of evaluated samples.
Together, these results validate EVA's evidence-grounded reasoning paradigm: tools are used sparingly on simple queries while iterative evidence acquisition is selectively applied when the initial evidence is insufficient.

\subsubsection{Performance--Latency Trade-off}

\begin{figure}[t]
    \centering
    \begin{minipage}{0.48\linewidth}
        \centering
        \includegraphics[width=\linewidth]{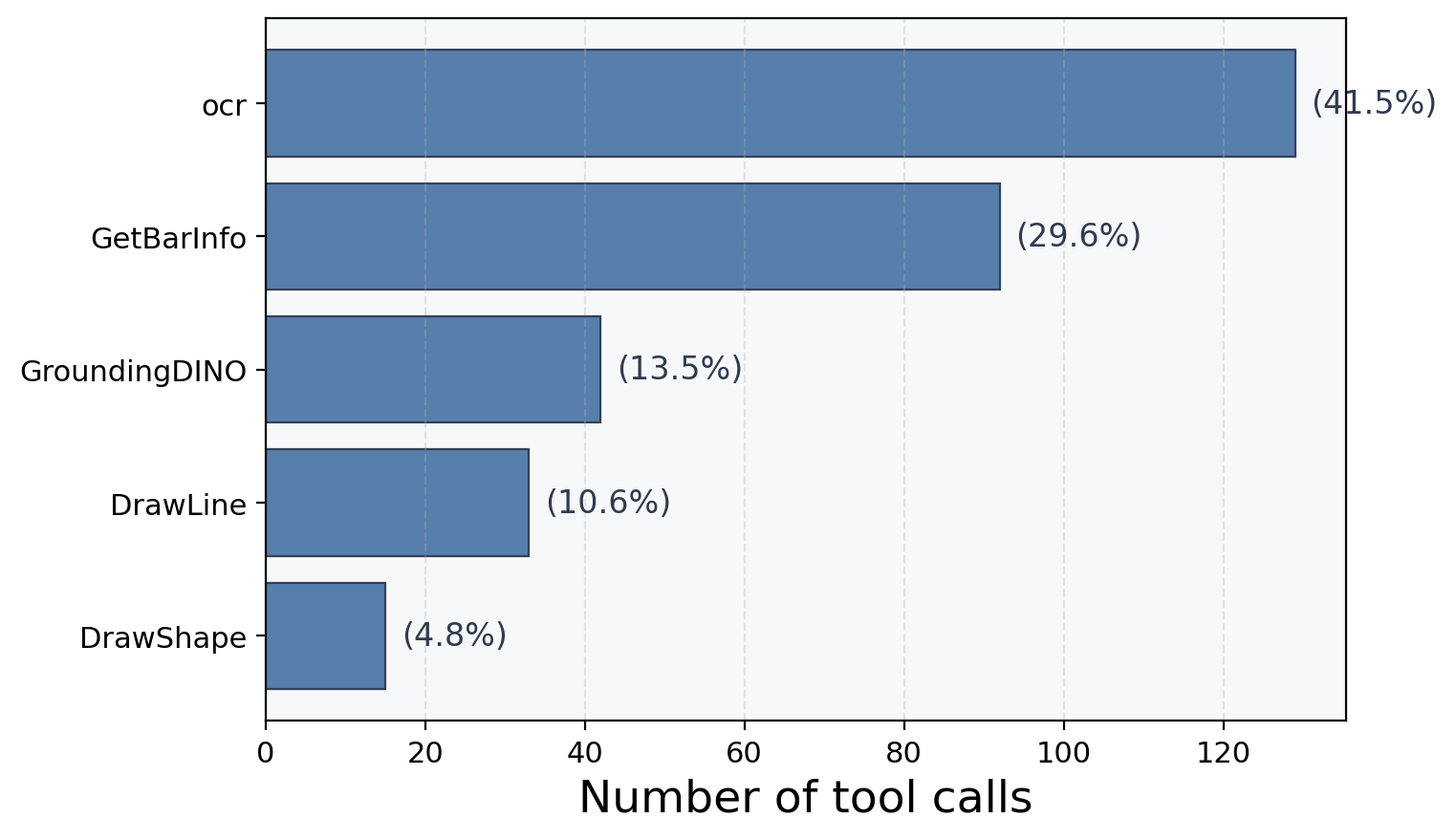}
        \caption{
        Distribution of external tool usage in EVA.
        OCR and chart-understanding tools dominate the invocation distribution, indicating that fine-grained perception remains a major source of multimodal hallucination.
        }
        \label{fig:tool_distribution}
    \end{minipage}
    \hfill
    \begin{minipage}{0.48\linewidth}
        \centering
        \includegraphics[width=\linewidth]{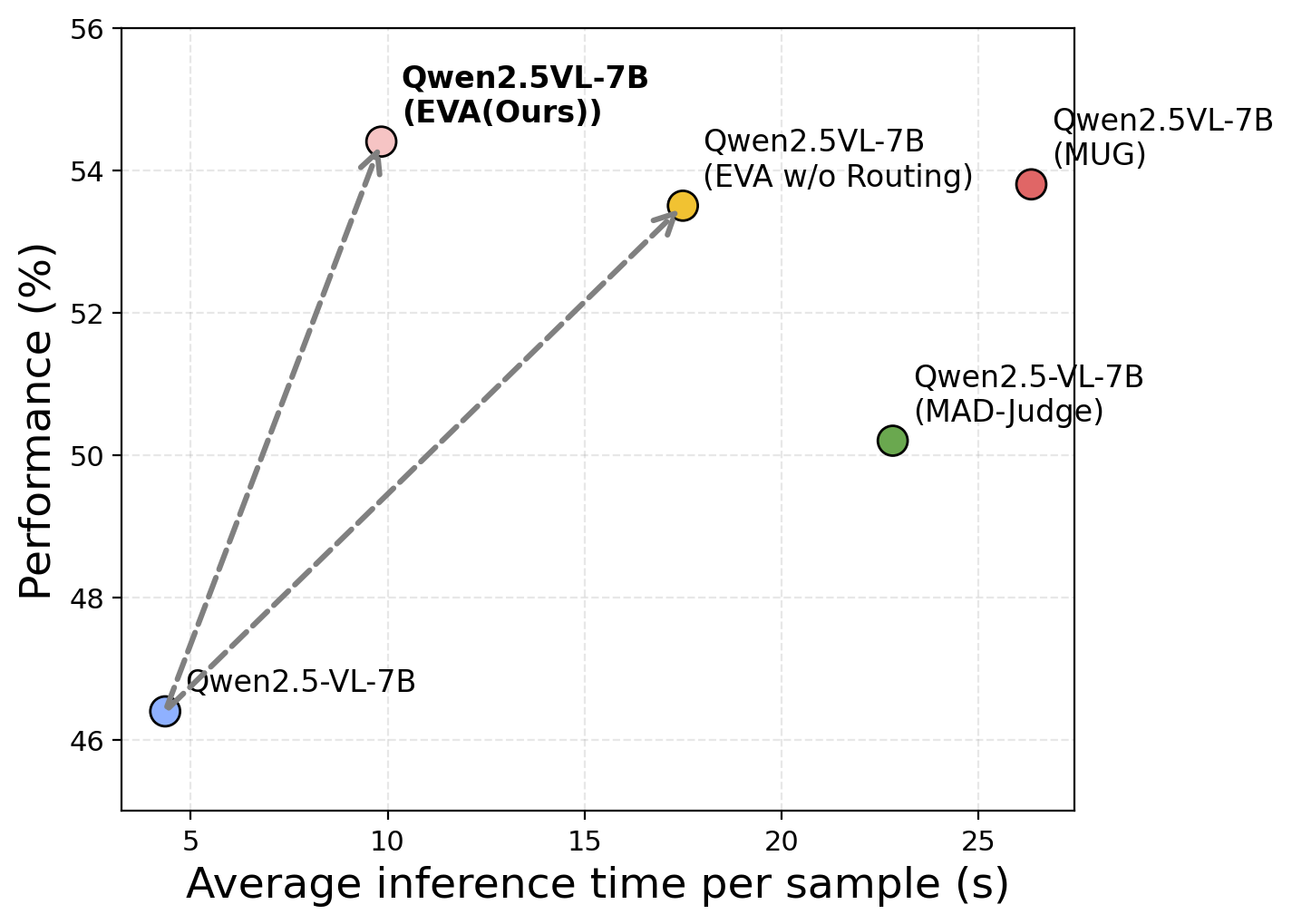}
        \caption{
        Performance--latency trade-off comparison.
        EVA achieves the best overall performance while maintaining significantly lower inference latency than multi-agent reasoning methods. w/o Routing: without Selective Evidence Routing.
        }
        \label{fig:latency_tradeoff}
    \end{minipage}
\end{figure}

Figure~\ref{fig:latency_tradeoff} compares EVA with direct inference and representative multi-agent reasoning methods under the performance--latency trade-off. All latency measurements are conducted on a single NVIDIA GeForce RTX 4090 GPU to ensure a fair and consistent comparison across methods.

Although multi-agent approaches such as MAD-Judge and MUG improve reasoning accuracy through iterative deliberation, they incur substantially higher inference costs due to repeated model interactions and unconditional reasoning expansion. In contrast, EVA achieves the best overall balance between reasoning performance and inference efficiency.
Several important observations can be drawn from the results:
(1) Direct inference with Qwen2.5-VL-7B achieves 46.4\% performance with low latency (4.35s), but suffers from severe hallucination and insufficient evidence verification.
(2) Multi-agent methods improve accuracy through iterative deliberation, with MAD-Judge reaching 50.2\% and MUG achieving 53.8\%. However, these gains come at significantly increased latency, requiring 22.83s and 26.35s per sample, respectively.
(3) EVA without routing already achieves strong reasoning performance (53.5\%) while reducing latency to 17.50s, demonstrating that explicit evidence acquisition and verification are more efficient than purely language-space deliberation. Full EVA further improves performance to 54.4\% while reducing latency to only 9.84s through \textit{Selective Evidence Routing}, which adaptively invokes external tools only when additional evidence is necessary.
These results suggest that explicit evidence-grounded reasoning provides a more efficient and scalable path toward hallucination-resistant multimodal reasoning than simply increasing language-space deliberation depth.

\subsubsection{Failure Source Analysis}
To better understand EVA's remaining limitations, we manually examine 50 failure cases and categorize each by the earliest decisive error in its reasoning process. Poor tool selection or incorrect tool parameters account for 46\% (23/50) of the failures, while incorrect or insufficient tool outputs account for another 22\% (11/50). The remaining 32\% (16/50) arise from incorrect interpretation or utilization of otherwise correct and sufficient evidence. In total, 68\% of the failures originate during evidence acquisition, whereas 32\% occur during evidence utilization. These results indicate that improving tool planning and evidence quality offers substantial headroom, while evidence-consistent answer generation remains important even when reliable observations have been obtained.

\subsection{Ablation Study}

\begin{table}[t]
\centering
\caption{Ablation study on HallusionBench. aAcc.: average accuracy, fAcc.: unique figure accuracy, qAcc.: unique question accuracy, Avg.: average score of the three metrics.}
\label{tab:ablation_hallusionbench}
\begin{tabular}{lcccc}
\toprule
{Method} & {aAcc.} & {fAcc.} & {qAcc.} & {Avg.} \\
\midrule
EVA & \textbf{70.4} & \textbf{45.4} & \textbf{47.3} & \textbf{54.4} \\
w/o Iterative Tool Replanning & 68.1 & 41.7 & 44.2 & 51.3 \\
w/o Evidence Consistency Verification & 67.4 & 40.9 & 43.5 & 50.6 \\
w/o Selective Evidence Routing & 69.5 & 44.3 & 46.7 & 53.5 \\
w/o Replanning \& Verification & 65.8 & 38.6 & 41.4 & 48.6 \\
\bottomrule
\end{tabular}
\end{table}

\paragraph{Component Ablation.}
To better understand the contribution of each component in EVA, we conduct ablation studies on HallusionBench. We progressively remove the proposed mechanisms from the full model, including \textit{Iterative Tool Replanning}, \textit{Evidence Consistency Verification}, and \textit{Selective Evidence Routing}. 
As shown in Table~\ref{tab:ablation_hallusionbench}, removing either \textit{Iterative Tool Replanning} or \textit{Evidence Consistency Verification} leads to clear performance degradation, demonstrating that both reliable evidence acquisition and evidence-consistent answer generation are critical for hallucination-resistant reasoning. In particular, removing \textit{Evidence Consistency Verification} causes a notable drop~(4.5\% in fAcc. and 3.8\% in qAcc.), suggesting that models may still generate unsupported conclusions even when external tool evidence is available. Similarly, removing \textit{Iterative Tool Replanning} weakens the model's ability to recover from incorrect tool choices or suboptimal tool parameters, resulting in lower overall accuracy.

When both replanning and verification are removed, the performance drops further to 48.6\% average accuracy, indicating that these two mechanisms are complementary. In contrast, removing \textit{Selective Evidence Routing} causes a smaller but consistent decrease. This suggests that routing mainly improves the efficiency--robustness trade-off by avoiding unnecessary tool usage for simple cases, while the main reliability gains come from the replan-and-verify execution pipeline. Overall, the ablation results validate the effectiveness of EVA's design: \textit{Iterative Tool Replanning} improves tool invocation reliability, \textit{Evidence Consistency Verification} mitigates evidence--reasoning inconsistency, and \textit{Selective Evidence Routing} provides adaptive and efficient evidence utilization.

\paragraph{Generalization to Frontier MLLMs.}
We further evaluate EVA on a stronger frontier MLLM by comparing GPT-5.5 with and without EVA under the same benchmark settings. As shown in Table~\ref{tab:eva_frontier}, EVA consistently improves performance across all four benchmarks. The largest improvement on HallusionBench suggests that explicit evidence acquisition and verification remain beneficial even when the backbone has stronger built-in visual understanding capabilities. A category-level analysis further indicates that these gains are particularly pronounced on visual-semantic reasoning problems involving charts and text-rich images, including a 4.2\% improvement on OCR-related questions. Notably, these improvements require only 0.14 external tool calls per sample on average, demonstrating that EVA can complement frontier MLLMs with limited additional tool usage.

\begin{table}[t]
\centering
\caption{Evaluation of EVA with a frontier MLLM. The same GPT-5.5 backbone is used for both settings. Tool calls are reported as per-sample averages.}
\label{tab:eva_frontier}
\small
\resizebox{0.8\linewidth}{!}{%
\begin{tabular}{lccccc}
\toprule
Method & MMMU & MMStar & HallusionBench & POPE & Tool calls \\
\midrule
GPT-5.5 (Direct) & 78.2 & 75.4 & 62.3 & 87.0 & 0 \\
GPT-5.5 + EVA & \textbf{79.3} & \textbf{76.2} & \textbf{64.4} & \textbf{89.6} & 0.14 \\
\bottomrule
\vspace{-1em}
\end{tabular}%
}
\end{table}

\paragraph{Controlled Tool-Use and Compute Comparison.}
We further investigate whether EVA's improvement can be explained solely by access to external tools or increased inference-time computation. Table~\ref{tab:eva_controlled_baselines} compares EVA against four training-free baselines with the same backbone, tool set, and evaluator. Despite having access to the same tools, OneShot and ReAct underperform EVA by 2.3 and 2.9 points, respectively, indicating that tool availability alone is insufficient. FixedSecondRound also remains 1.9 points below EVA, despite requiring more model calls, generated tokens, tool calls, and latency. More importantly, SameEvidenceSelfRefine receives exactly the same acquired evidence as EVA and uses the same average number of model and tool calls, yet scores 1.3 points lower while generating more tokens and incurring higher latency. Together, these controlled comparisons suggest that EVA's improvements stem from its coordinated selective routing, conditional replanning, and evidence consistency verification, rather than simply from additional tool access, evidence availability, or inference computation.

\begin{table}[t]
\centering
\caption{Controlled comparison on HallusionBench using Qwen2.5-VL-7B. All methods are training-free and share the same tool set and evaluator. Resource statistics are per-sample averages.}
\label{tab:eva_controlled_baselines}
\small
\resizebox{\linewidth}{!}{%
\begin{tabular}{lccccc}
\toprule
Method & HallusionBench Avg. & Model calls & Generated tokens & Tool calls & Latency (s) \\
\midrule
OneShot & 52.1 & 1.92 & 175.80 & 1.06 & 7.98 \\
ReAct & 51.5 & 2.18 & 183.54 & 1.18 & 8.72 \\
FixedSecondRound & 52.5 & 2.84 & 242.80 & 1.98 & 13.53 \\
SameEvidenceSelfRefine & 53.1 & 2.62 & 221.38 & 0.32 & 10.20 \\
EVA & \textbf{54.4} & 2.62 & 218.26 & 0.32 & 9.84 \\
\bottomrule
\end{tabular}%
}
\end{table}

\section{Conclusion}

We presented EVA, an evidence-seeking visual agent for reliable multimodal reasoning. EVA improves reasoning reliability through three key mechanisms: \textit{Selective Evidence Routing}, \textit{Iterative Tool Replanning}, and \textit{Evidence Consistency Verification}. Together, these components enable EVA to adaptively acquire external evidence, refine tool usage, and ensure that final predictions remain grounded in verifiable observations.
Extensive experiments show that EVA consistently reduces hallucination and improves multimodal reasoning performance across challenging benchmarks, outperforming direct MLLM inference and existing multi-agent reasoning methods while avoiding unnecessary tool usage. Our results highlight the importance of explicit evidence acquisition and verification for building reliable multimodal agents.

\section*{Acknowledgement}
This work was supported by the grants from the National Natural Science Foundation of China 62372014, Beijing Nova Program and Beijing Natural Science Foundation 4252040.

\bibliographystyle{plainnat}
\bibliography{ref}

\end{document}